\documentclass{article}
\usepackage{spconf,amsmath,graphicx}
\usepackage{multirow}
\usepackage{hyperref}
\usepackage{comment}
\usepackage{bm}
\usepackage{cite}
\title{GPS-Denied Navigation using SAR Images and Neural Networks}
\name{Teresa White$^1$, Jesse Wheeler$^2$, Colton Lindstrom$^3$, Randall Christensen$^{3*}$, Kevin R. Moon$^{1*}$\thanks{$^*$Equal contributions. This research was partially supported by Sandia National Labs on grants 201782, 202136, and 202854.}}
\address{ $^1$Dept. Mathematics \& Statistics, $^3$Dept. Electrical \& Computer Engineering\textemdash Utah State University \\
    $^2$Dept. Statistics\textemdash University of Michigan}
\begin{document}
\ninept
\maketitle
\begin{abstract}

Unmanned aerial vehicles (UAV) often rely on GPS for navigation. GPS signals, however, are very low in power and easily jammed or otherwise disrupted. This paper presents a method for determining the navigation errors present at the beginning of a GPS-denied period utilizing data from a synthetic aperture radar (SAR) system. This is accomplished by comparing an online-generated SAR image with a reference image obtained apriori. The distortions relative to the reference image are learned and exploited with a convolutional neural network to recover the initial navigational errors, which can be used to recover the true flight trajectory throughout the synthetic aperture. The proposed neural network approach is able to learn to predict the initial errors on both simulated and real SAR image data.

\end{abstract}

\begin{keywords}
SAR images, CNNs, GPS-denied navigation
\end{keywords}

\section{Introduction}

Unmanned aerial vehicles (UAV) have many applications such as search and rescue operations, military applications, agricultural applications, surveillance and more. In many of these applications, knowledge of the current flight trajectory of the UAV is necessary to carry out the UAV’s mission. In normal circumstances, GPS is typically used to obtain this knowledge. However, in abnormal and important circumstances, such as navigating in a hostile airspace, the GPS signal may be denied or otherwise degraded. Here, we propose a deep neural network model that accurately estimates the flight trajectory by comparing a synthetic aperture radar (SAR) image obtained online to a previously-obtained reference image.

%\subsection{Motivation}
GPS signal vulnerability to many forms of interference has motivated research in alternative approaches to localization~\cite{raquet_non-gnss_2008,veth_navigation_2011}. Radar is one of many active approaches, which, in contrast to image-based methods, is independent of lighting conditions, operates in any weather condition, and provides an accurate measurement of range. Many approaches have been developed which exploit artifacts in SAR imagery to determine navigation errors, either relative to the beginning of the synthetic aperture~\cite{quist_radar_2016,quist_radar_2016-1,scannapieco_experimental_2018,samczynski_coherent_2010,samczynski_superconvergent_2012} or in a global/absolute reference frame~\cite{hostetler_nonlinear_1983,hollowell_heli/sitan:_1990,bergman_terrain_1999,nordlund_marginalized_2009,kim_terrain-referenced_2018,sjanic_simultaneous_2010,greco_sar-based_2012,nitti_feasibility_2015}.

%\subsection{Literature Review}
Back-projection-based SAR imaging is a technique for producing radar  images during arbitrary flight trajectories~\cite{zaugg_generalized_2015,carrara_spotlight_1995,ulaby_microwave_2014}. SAR images are created by sending a series of radar pulses during a portion of the vehicle's trajectory. The back-projection algorithm (BPA) is then used to synthesize the reflected signals into a two or three-dimensional SAR image. To obtain an accurate image, back-projection requires an accurate estimate of the position of the radar antenna throughout the synthetic aperture. If the assumed flight trajectory is inaccurate, the resulting SAR image will be distorted by shifts and blurs in the along track (AT) and cross track directions (CT)~\cite{christensen2019toward}. In a typical SAR application, an inertial navigation system utilizes GPS measurements to generate the best estimate of the aircraft trajectory and minimize SAR image distortions. In contrast, we focus on the reverse problem, where the distortions in the SAR image are exploited to  estimate the vehicle trajectory throughout the synthetic aperture.  To reduce the dimensionality of the estimation problem, we use the well-known error dynamics for inertial navigation systems, effectively reducing the number of solve-for parameters to nine initial errors.

For the simple case of constant horizontal position errors, identification of SAR image shifts with respect to an \textit{a priori} SAR map provide easily exploitable information regarding   position errors. When velocity and attitude errors are included, however, the mapping from initial errors to image distortions is more complex~\cite{christensen2019toward,lindstrom2020sensitivity}.

There is a strong precedent for applying machine learning techniques  to SAR imagery. The most common application is automatic target recognition~\cite{berisha2007sparse,chen2014,chen2016,wilmanski2016,zhao2001,krishnapuram2005semi}. Other applications include the use of deep learning to detect changes in radar images of the same landscape over time \cite{gong2016} or auto-focusing blurry radar images \cite{mason2017}. These latter two applications focus primarily on image reconstruction. Despite a wide range of target recognition problems, there is no precedent in the literature to use machine learning to predict initial navigational errors in a regression-based framework, further motivating the need for our analysis.

Our problem shares some similarities with the task of image registration~\cite{hero2002applications,tipping2003bayesian,pickup2007bayesian,wyawahare2009image} in that a reference image and an input image are both considered. However, in image registration, the goal is to align the input image with the reference image. In our setting, we extract information (the true trajectory of the aircraft) from the degree to which the input image and reference image do not match, as well as the manner in which they differ.

%\subsection{Contributions}
We show that neural networks can be used to develop a black-box model of the distortion dynamics that takes as input the distorted image and a reference image and outputs the true flight trajectory. In particular, this work shows that a modern convolutional neural network (CNN) architecture produces the best results in estimating the flight trajectory using both simulated and real data. %It is demonstrated using a specific approach for characterizing the distortions in the input image: stacking the reference image with the distorted input image so that it is treated as a different channel in the first convolutional layer. 

\section{Background}
In strapdown inertial navigation systems, measurements of specific force $\boldsymbol{\nu^b}$ and angular rate $\boldsymbol{\omega^b}$ are used to propagate the position $\boldsymbol{p}^n$, velocity $\boldsymbol{v}^n$, and attitude quaternion $q_b^n$ of the aircraft, via the following differential equations
\begin{equation}
\left[\begin{array}{c}
\dot{\boldsymbol{p}}^{n}\\
\dot{\boldsymbol{v}}^{n}\\
\dot{q}_{b}^{n}
\end{array}\right]=\left[\begin{array}{c}
\boldsymbol{v}^{n}\\
T_{b}^{n}\boldsymbol{\nu}^{b}+\boldsymbol{g}^{n}\\
\frac{1}{2}q_{b}^{n}\otimes\left[\begin{array}{c}
0\\
\boldsymbol{\omega}^{b}
\end{array}\right]
\end{array}\right],\label{eq:cov-veh2}
\end{equation}
where $b$ represents a coordinate system attached to the body and $n$ represents a locally-level frame with the x-axis aligned with the nominal vehicle velocity, the z-axis in the direction of gravity, and the y-axis determined by the right-hand-rule.  Thus the $n$ frame represents an along-track, cross-track, down coordinate system.

At regular intervals, an external aiding device, such as GPS, provides information to correct the navigation states, typically in the framework of an extended Kalman filter.  In between aiding device measurements, or in the absence thereof, the navigation errors accumulate over time.  For straight and level flight, typical of SAR data collection, the growth of error is modeled to first order:
\begin{equation}\label{eq:nav-error-prop}
\left[\begin{array}{c}
    \delta\boldsymbol{p}^n\left(t\right)\\
    \delta\boldsymbol{v}^n\left(t\right)\\
    \delta\boldsymbol{\theta}^n\left(t\right)\\
\end{array}\right] 
    = \Phi\left(t,t_0\right)
\left[\begin{array}{c}
    \delta\boldsymbol{p}^n\left(t_0\right)\\
    \delta\boldsymbol{v}^n\left(t_0)\right)\\
    \delta\boldsymbol{\theta}^n\left(t_0\right)\\
\end{array}\right], 
\end{equation}
where $\Phi\left(t,t_0\right)$ is the state transition matrix defined as
\begin{equation}
\Phi\left(t,t_0\right)=
    \left[\begin{array}{ccc}
I_{3\times3} & I_{3\times3}\Delta t & \left(\boldsymbol{\nu}^{n}\times\right)\frac{\Delta t^{2}}{2!}\\
0_{3\times3} & I_{3\times3} & \left(\boldsymbol{\nu}^{n}\right)\times\Delta t\\
0_{3\times3} & 0_{3\times3} & I_{3\times3}
\end{array}\right].
\end{equation}

The relationship between the estimated, true, and error parameters is defined by the additive or multiplicative relationships
\begin{equation}\label{eq:nav-pos-correction}
\boldsymbol{p}^{n}\left(t\right) = \hat{\boldsymbol{p}}^{n}\left(t\right) + \delta\boldsymbol{p}^{n}\left(t\right)
\end{equation}
\begin{equation}\label{eq:nav-vel-correction}
\boldsymbol{v}^{n}\left(t\right) = \hat{\boldsymbol{v}}^{n}\left(t\right) + \delta\boldsymbol{v}^{n}\left(t\right)
\end{equation}
\begin{equation}\label{eq:nav-att-correction}
q_{b}^{n}\left(t\right)  = \left[\begin{array}{c}
1\\
-\frac{1}{2}\delta\boldsymbol{\theta}^{n}\left(t\right) 
\end{array}\right]\otimes\hat{q}_{b}^{n}\left(t\right). 
\end{equation}
Thus given an initial condition for the navigation errors, the true navigation states can be recovered by propagating the errors using \autoref{eq:nav-error-prop} and substituting the result into Equations \ref{eq:nav-pos-correction}-\ref{eq:nav-att-correction}.  The task of correcting errors in the estimated trajectory is, therefore, reduced to determining the nine errors at the beginning of the time interval of interest.  A more detailed discussion of the inertial navigation framework and error modeling is provided in~\cite{lindstrom2020sensitivity,christensen2019toward}.

Previous sensitivity studies discovered the complex and seemingly ambiguous relationship between the nine initial navigation errors and the induced SAR image distortions, summarized in~\autoref{tbl:sensitivity}~\cite{christensen2019toward}. In practical navigation systems, however, the down component of position and velocity errors are easily controlled with a barometric altimeter. The absence of vertical errors, combined with relatively small sensitivity to typical attitude errors, results in SAR image distortions dominated by position and velocity errors in the AT and CT directions.
\begin{table}[!ht]
\centering
\scalebox{0.8}{
\begin{tabular}{|c|c|c|}\hline
Error &  Shift Direction & Blur Direction \\\hline
AT Position &  AT     & None \\\hline
CT Position &  CT     & None \\\hline
D Position  &  CT     & None \\\hline
AT Velocity &  None   & AT   \\\hline
CT Velocity &  AT     & None \\\hline
D Velocity  &  AT     & None \\\hline
AT Attitude &  None   & Small AT \\\hline
CT Attitude &  None   & Small AT \\\hline
D Attitude  &  None   & Small AT \\\hline
\end{tabular}
}
\caption{Effect of navigation errors on BPA-SAR images.}
\label{tbl:sensitivity}
\end{table}

\begin{figure}
    \centering
    \includegraphics[width=\columnwidth]{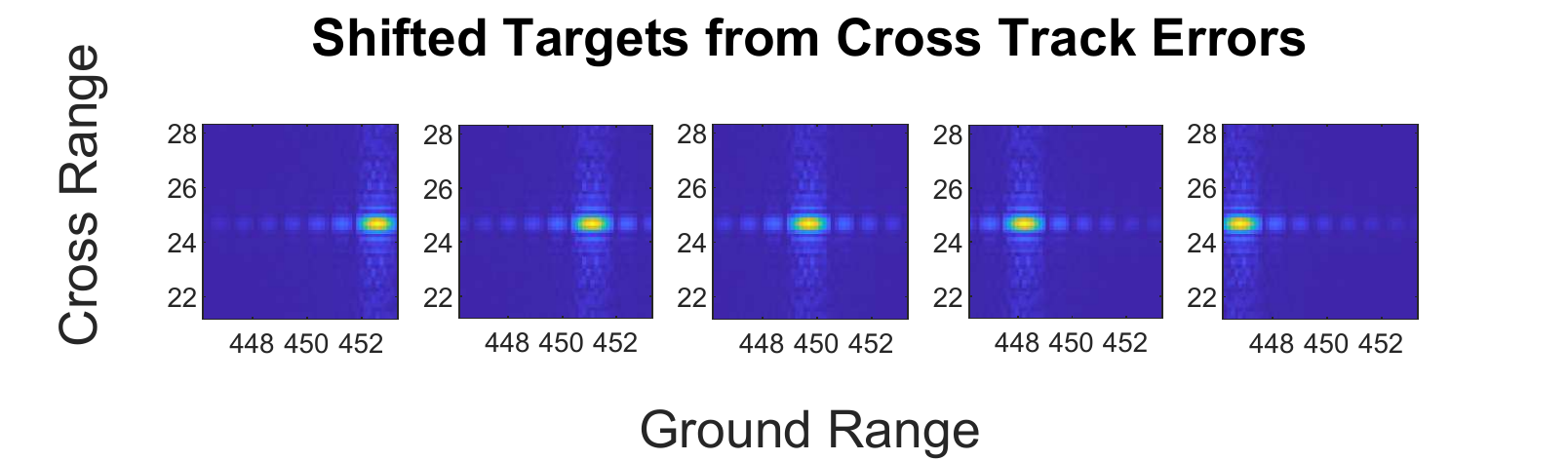}
    \includegraphics[width=\columnwidth]{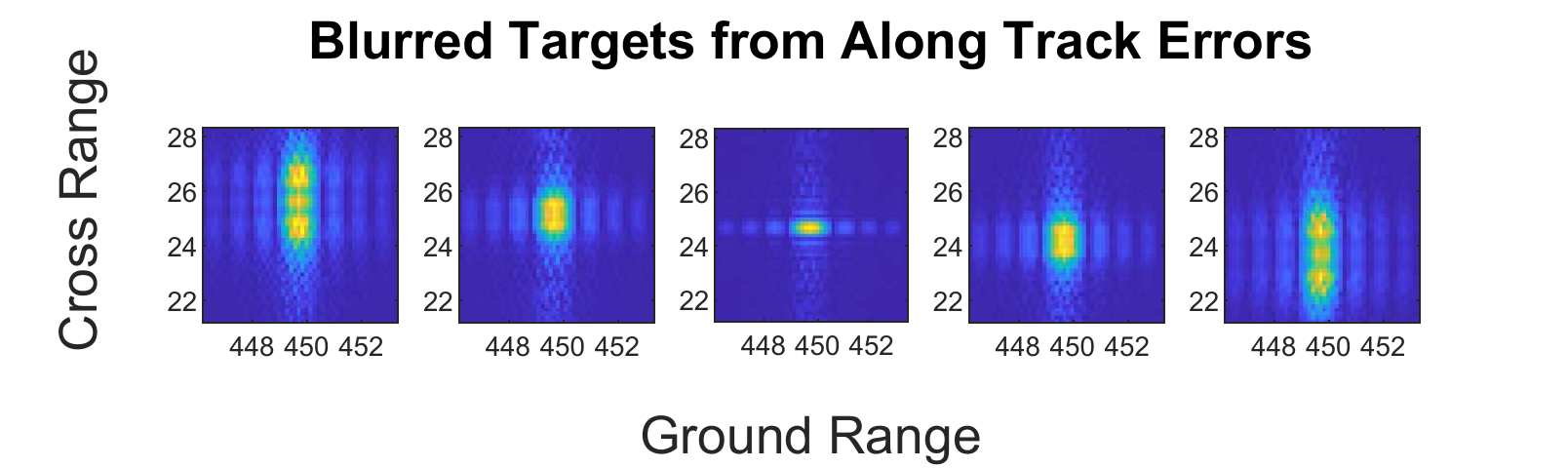}
    \caption{Demonstration of shifting and blurring distortions due to navigation errors. 
        (Top) Shift distortions caused by errors in cross track position. From left to
        right, cross track errors are -3, -1.5, 0, 1.5, and 3 meters. (Bottom) Blur 
        distortions caused by errors in along track velocity. From left to right, along
        track errors are -0.4, -0.2, 0, 0.2, and 0.4 meters per second.}
    \label{fig:Errors}
\end{figure}

While image distortions are characterized by shifts and blurs in different directions (see~\autoref{fig:Errors}), certain combinations of navigation errors can create image distortions that appear similar to the human eye, making it difficult to determine the true source of distortion and thus the true trajectory error. We hypothesize that different error combinations create subtly different image distortions that can be learned using neural networks to predict the initial navigation errors.

\section{The SAR Data}\label{data}

%Three different SAR image data were generated to apply our CNN architecture and estimate their correct flight trajectories. Each of these data has a different synthetic aperture length. The first data is the simulated data with a synthetic aperture length of 5 seconds. The second data is the real-2-sec data with a synthetic aperture length of 2 seconds. Then, the third data is the real-10sec data with a synthetic aperture length of 10 seconds. For each of them, six different datasets were produced. 

Three different sets of SAR image data were generated for training and testing: simulated data with a 5 second aperture length (sim-5-sec), and two real datasets with 2 (real-2-sec) and 10 second (real-10-sec) aperture lengths, respectively. Simulated data is generated from flight and radar software developed in MATLAB. The software has the capability to generate a flight trajectory, populate the ground with radar targets, and form both truth and distorted SAR images via the BPA from synthesized radar data~\cite{duersch_analysis_2015,lindstrom2020sensitivity}.  For the real datasets, the radar data is obtained from flight tests of the Space Dynamics Laboratory X-band radar system.  A sub-centimeter-accurate, post-processed trajectory is used as truth, and corrupted to synthesize distorted SAR images.

% \usepackage{multirow}

\begin{table}[!ht]
\centering
\scalebox{0.8}{
\begin{tabular}{|c|c|c|c|c|c|c|}\hline
Scenario \# &  AT Pos & CT Pos & D Pos & AT Vel & CT Vel & D Vel   \\\hline
{1} &  x     & x     &       &       &       &         \\\hline
{2} &        &       &       & x     & x     &         \\\hline
{3} &  x     & x     &       & x     & x     &         \\\hline
{4} &  x     & x     & x     &       &       &          \\\hline
{5} &        &       &       & x     & x     & x        \\\hline
{6} &  x     & x     & x     & x     & x     & x        \\\hline
\end{tabular}
}
\caption{Summary of scenarios and corresponding initial errors.  %The network attempted to model both of these error states. 
%In Dataset 7, however, North, East, and Down Attitude errors were also present in the generated data, but the Neural Network only tried to predict North Position and East Position. 
}
\label{datasets}
\end{table}

%\subsection{Splitting Data}

For each of the simulated and real datasets, six different scenarios were studied, with varying numbers and combinations of active navigation errors, as shown in \autoref{datasets}. For each of the six scenarios, a total of 192, 185, and 175 unique target locations were considered for the sim-5-sec, real-2-sec, and real-10-sec dataset, respectively. For each target, we generated 100 distorted images of size $80 \times 80$ pixels, paired with the corresponding navigation errors. The datasets were then sorted by targets into training, validation, and testing sets as shown in \autoref{splitting_data}. We train on a set of targets, and validate and test on an entirely different set of targets. Thus a network's performance on the test dat reflects its ability to generalize to new locations. 

\begin{table}[!ht]
\centering
\scalebox{0.77}{
\begin{tabular}{|c|c|c|c|c|c|c|}\hline
\multirow{2}{*}{Dataset} &  \multicolumn{2}{|c|}{ Training }       &  \multicolumn{2}{|c|}{Validation}    &  \multicolumn{2}{|c|}{Testing}\\
                             &  \# targets   &  \# images              &    \# targets  &  \# images          &    \# targets   &  \# images    \\\hline
Sim-5-sec               & 134 &   13500                 &  19  &     1900            &  39   &   3800      \\\hline
Real-2-sec               & 130 &   13000                 &  18  &     1800            &  37   &   3700      \\\hline 
Real-10-sec              & 122 &   12300                 &  17  &     1700            &  36   &   3500      \\\hline
\end{tabular}
}
\caption{Data split details. The six datasets of the sim-5-sec, real-2-sec, and real-10-sec data were approximately split into a $70\%$ training, $10\%$ validation, and $20\%$ testing set. %As shown, for each of the six datasets of the real-2-sec data, the training set had 130 unique target locations and 13000 distorted images.
}
\label{splitting_data}
\end{table}

%\subsubsection{Simulated Data with an aperture length of 5 seconds}
%For each of the nine different datasets for the simulated with an aperture length of 5 seconds, a total of 192 unique target locations were considered and 100 distorted images paired with initial navigational errors were created for each target. Then, the data were sorted into training (134 targets), validation (19 targets), and testing (39 targets) datasets.

%\subsubsection{Real Data with an aperture length of 2 seconds}
%In a similar way, for each of the nine different datasets for this real data, a total of 185 unique target locations were considered and 100 distorted images were created for each target. We split the data into training (130 targets), validation (18 targets), and testing (37 targets) datasets.

%\subsubsection{Real Data with an aperture length of 10 seconds}
%For each of the nine different datasets for this real data, a total of 175 unique target locations were considered and 100 distorted images were created for each target. We split the data into training (122 targets), validation (17 targets), and testing (36 targets) datasets.

%Sorting the data this way allows us to train on a set of targets, and validate and test on an entirely different set of targets. This demonstrates the network's ability to generalize to new locations. 

%\subsection{Data Preprocessing}

%SAR images measure the intensity of the recieved signal which is transmitted from the aircraft and reflects off of objects in the environment. These intensities are inherently... 

The brightness of each pixel of a SAR image represents the reflectivity of all targets within that pixel's region.  The intensities of the pixels are inherently right-skewed and subject to changes in magnitude based on the geometric conditions. Thus, we preprocessed the images and errors as follows.

Let $X$ be an input image, $X'$ the corresponding reference image, and $y_{\alpha_\beta}$ the corresponding navigational error vector. We first log-transform each pixel in each image (including the reference images): $L=\log_{10}(X)$ as is common practice in SAR images. We then standardize each pixel in the image to ensure standardized inputs to the neural network by subtracting the mean and dividing by the standard deviation: $Z=\frac{L-\mu_L}{\sigma_L}$, where $\mu_L$ and $\sigma_L$ are estimated on an image-by-image basis. 

Each of the navigational errors $y_{\alpha_{\beta}}$ are measured on different scales. We therefore standardize the navigational errors as $\bm{s} = \frac{\bm{y} - \mu_{\bm{y}}}{\sigma_{\bm{y}}}$ to ensure equal consideration of all navigational errors during training.
We estimate $\mu_{\bm{y}}$ and $\sigma_{\bm{y}}$ from the entire dataset.
In practice, the standardization of the navigation parameters would use a mean of 0 (assuming unbiased navigation errors on average) and a standard deviation estimated from a extended Kalman filter \cite{kalman1960new}.

%To compare the input image with the reference image, we stack the normalized input, the reference image, and the difference image between the normalized input and reference image, i.e. $D=L-L'$. Then it is used as input in the deep network model.

%Because of this transformation, the Mean Squared Error (MSE) for any prediction can be compared against the value of 1: if the network guesses that the initial error state is 0 or randomly guesses with a mean 0 and a similar standard deviation, then the network will have an estimated MSE of 1.  If the network is accurately estimating the relationship between the reference image, distorted image, and initial navigational error states, then the MSE would be less than 1. Thus this serves as a benchmark for learning where an MSE less than one indicates that the neural network is learning relevant information for this task.

\section{Neural Network Methods}

Convolutional neural networks (CNN) are a special case of a neural network that achieve very good performance on image recognition tasks~\cite{ResNet,krizhevsky2012imagenet,sainath2013deep}. We therefore use CNN architectures to learn the initial navigational errors. We considered multiple ResNet~\cite{ResNet} and Wide ResNet~\cite{WideResNet} architectures. Each of these architectures is built with the residual block. This block allows networks to be much deeper (more layers and hence better at dealing with images) while simultaneously mitigating the issue of the vanishing gradient \cite{ResNet}. 

%The residual block  is still considered a state-of-the-art way of dealing with image data. 

%Originally, the application of simple CNN architectures on the simulated dataset gave good results. However, when we applied them to the real data, these simple networks did not have the desired performance. For this reason, more advanced network architectures were considered. Below is a list of all modern CNN architectures that have been considered for this model: 

%\begin{itemize}
%    \item ResNet 18 \cite{ResNet}
%    \item ResNet 34 \cite{ResNet}
%    \item ResNet 50 \cite{ResNet}
%    \item ResNet 101 \cite{ResNet}
%    \item ResNet 152 \cite{ResNet}
%    \item Wide ResNet 50\_2 \cite{WideResNet}
%    \item Wide ResNet 101\_2 \cite{WideResNet}
%\end{itemize}

%Each of these architectures is built with the same basic idea: the residual block. This block allows networks to be much deeper (more layers and hence better at dealing with images) while simultaneously mitigating the issue of the vanishing gradient \cite{ResNet}. The residual block technique is still considered a state-of-the-art way of dealing with image data. 

Each of the ResNet and Wide ResNet architectures  have pre-trained models trained on large image datasets. We settled on the Wide ResNet 50\_2 network because it outperformed all other models, and because its wide (and relatively shallow) nature enables faster training. Our input consists of a 3-channel image with the distorted image, the reference image, and the difference image as each of the channels. This is fed into a randomly initialized convolutional layer followed by the ResNet architecture. The final layer of ResNet is replaced with a  fully connected layer with the same number of outputs as error states. In our experiments, we found that fine-tuning all of the ResNet weights resulted in better performance than ``freezing" any of the weights or randomly initializing them. This is remarkable considering the datasets used for pretraining contained natural images, which differ considerably from SAR images. This suggests that the pretrained network has learned  image features (e.g. lines and other shapes) that are also present in SAR images. 

%In the initial attempts at estimating initial navigational errors, it was explored whether or not starting with pre-trained weights or randomly initializing the model's weights would produce the best results. Unsurprisingly, in all cases, the model with pre-trained weights outperformed models that had weights that were trained using only the reference chips and distorted images. It was further discovered that starting with pre-trained weights but then allowing these weights to be modified specifically for the problem at hand produced better results than ``freezing" the already pre-trained weights. 

%More testing suggested that all of the architectures that were listed above had similar performance; there was no single architecture that consistently outperformed the others. It was decided to use Wide ResNet 50\_2 because it outperformed the other models more frequently than any other, and because the wide (and relatively shallow) nature of the Wide ResNet 50\_2 model makes it faster to train, and therefore more time can be devoted to trying specific hyper-parameters. 

%\subsection{Transfer Learning}

We considered two main approaches for training for the real data. In the first, we simply trained on the real data using the split described in \autoref{splitting_data}. In the second, we used transfer learning. Transfer learning is a  technique that can help to improve the learning in a new task through the transfer of knowledge from another related task. For example, using the pretrained Wide ResNet 50\_2  is a form of transfer learning.  The goal of using transfer learning is to fully learn the target task given the transferred knowledge instead of learning from scratch. It can often speed up the training and improve the performance of the model. To do transfer learning in this approach, we pretrained the network (which was already pretrained on other image data) on the simulated training data and then trained  on the real training data. The goal of this study was to determine if training on simulated data can improve the performance on real data.
%The dataset used to test this approach is dataset \#2 where N Vel and E Vel are the only errors present, and the model attempts to predict both error states. 

We used the average mean squared error (MSE) loss function across all considered initial errors for training and for testing:
\[\text{MSE} = \frac{1}{mn}\sum_{\alpha=1}^n\sum_{\beta=1}^m \left(s_{\alpha_\beta} - \hat{s}_{\alpha_\beta}\right)^2, \]
where $m$ is the number of error states considered, $n$ is the number of training images in the data, $s_{\alpha_\beta}$ is the true standardized error of the $\beta$th error and the $\alpha$th image, and $\hat{s}_{\alpha_\beta}$ is the corresponding error estimate by the neural network.
The MSE for any prediction can be compared against the value of 1: if the network guesses that the initial error state is 0 or randomly guesses with a mean 0 and a similar standard deviation, then the network will have an estimated MSE of 1.  Thus this serves as a benchmark for learning, where an MSE less than one indicates that the neural network is learning relevant information for this task.

\section{Experimental Results}

%All the results shown in this section were obtained using the two methods from above.

%\subsection{CNN Method}\label{sim_real_data}

\autoref{sim_real_results} lists the results obtained in all six scenarios for both the sim-5-sec and the real-2-sec dataset. For all scenarios in the simulated dataset, the MSE is less than $1$ for all considered errors. In particular, the simulated dataset performs well for scenarios 1 and 2, where no ambiguity exists.  These results suggest that the neural network is capable of characterizing both shifts and blurs, in the absence of ambiguity. However, as more errors are introduced, the performance degrades, suggesting difficulties in resolving ambiguous error sources. 

%Adding the accuracy plot Dataset 1
\begin{figure}[htb]
    \centering
    \includegraphics[width=0.28\textwidth]{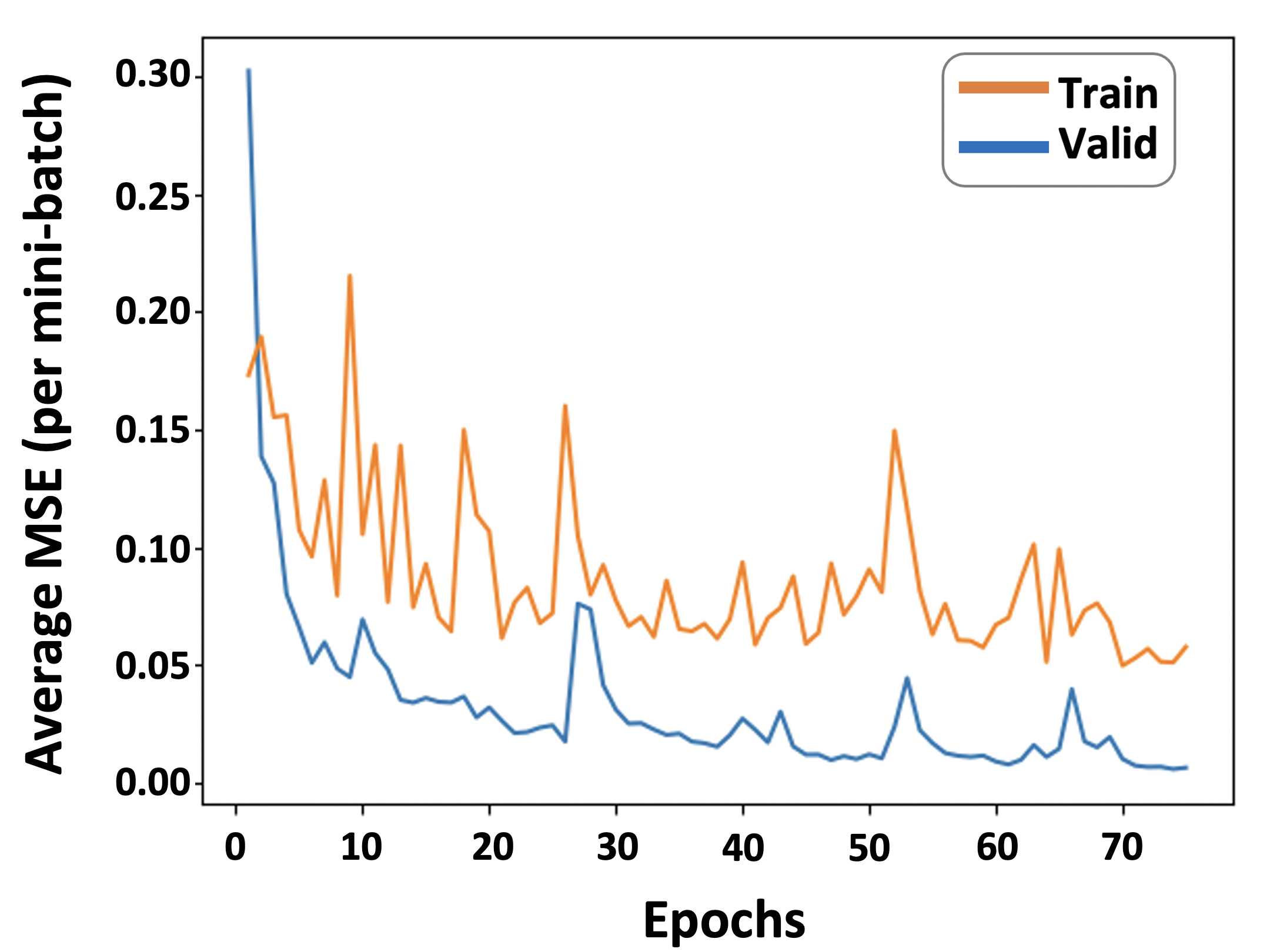}
    \caption{Training and validation MSE as a function of training epoch for scenario 1. The gap between the errors is small.}
    \label{fig:Accuracy_training_Data1}
\end{figure}

%(A paragraph on our test vs. validation vs. training error)

\begin{figure}[htb]
    \centering
    \includegraphics[width = .45\textwidth]{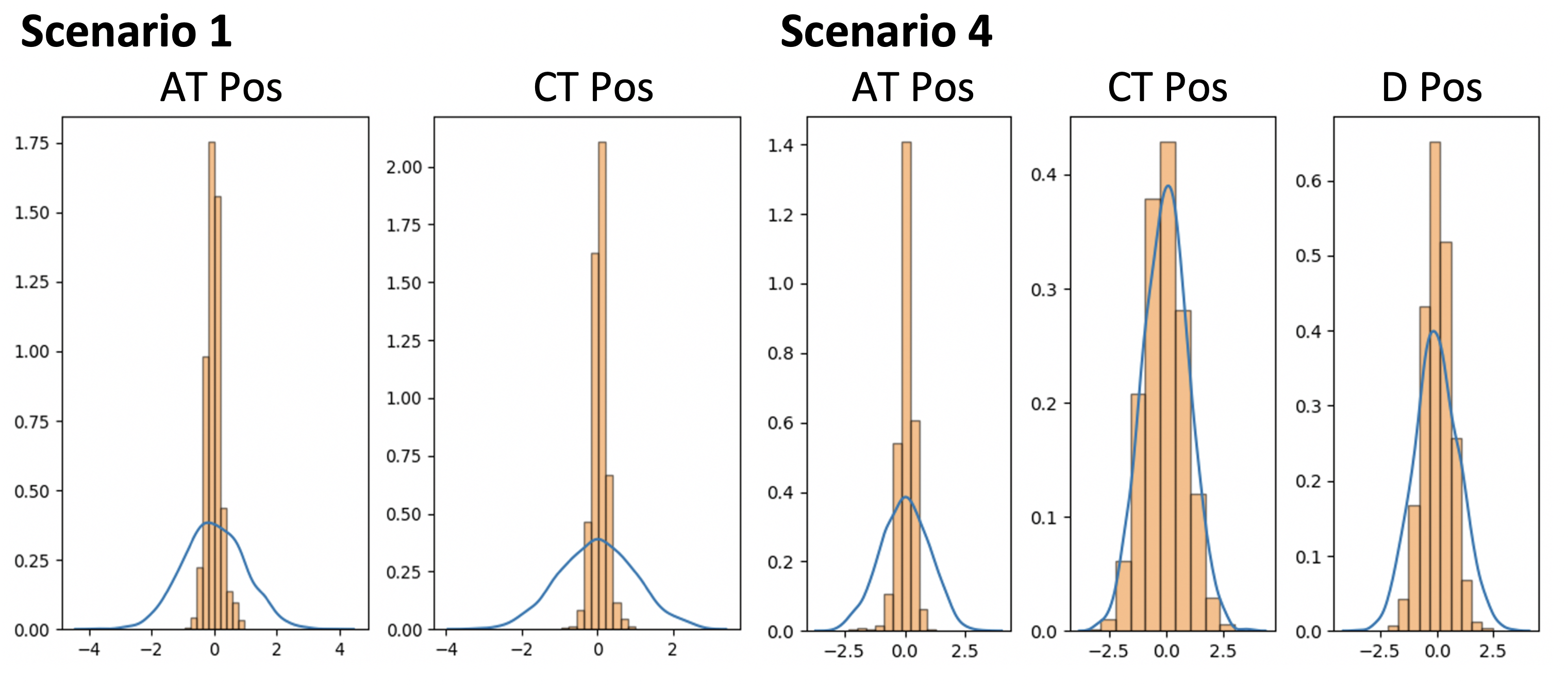}
    \caption{Distribution of Error States before (blue line) and after (histogram) estimation for the real-2-sec dataset for scenarios 1 and 4. In both cases, the error distributions are more concentrated around zero, indicating that the network has improved our ability to estimate  navigation errors. No outliers have been introduced by the network.}
    \label{fig:Data4Errors}
\end{figure}

In the real dataset, the MSE is below 1 for all errors in scenarios 1 and 4. The network performs particularly well in scenario 1 with a low MSE (see also \autoref{fig:Accuracy_training_Data1}). It is noteworthy that for scenario 4, ambiguity exists in CT shifts, due to either CT or D position errors. The network, however, is able to identify subtleties in the distorted SAR images and properly attribute the effect to the corresponding error.  \autoref{fig:Data4Errors} shows the distribution of the error states before and after estimation of the navigation errors for scenarios 1 and 4. The resulting error distributions are more concentrated around zero, indicating that the neural network is able to improve our knowledge of the initial error states. Furthermore, no outliers are introduced by the neural network. 
However, as was the case for the simulated dataset, increasing numbers of ambiguous error sources results in degraded overall performance, with an MSE near 1.0 in cases where little information is learned by the network.

%As we can see, the MSE for the dataset \#4 of the real-2-sec data is less than $1$ for all of the errors present in that dataset. It indicates that our network is predicting the N Pos, E Pos and D Pos error states well.

%The Wide ResNet 50\_2 architecture was applied on each of the six datasets of the simulated and real-2-sec data. 

%A possible table

\begin{table}[ht]
\centering
\scalebox{0.7}{
\begin{tabular}{|c|l|c|c|c|c|c|c|}\hline
Scenario \#         & Dataset               & AT Pos  & CT Pos  & D Pos  &  AT Vel & CT Vel  & D Vel  \\\hline
\multirow{2}{*}{1} & MSE (Sim-5-sec)  & 0.0594 & 0.0289 & N/A    &  N/A   &  N/A   & N/A  \\
                   & MSE (Real-2-sec) & 0.0563 & 0.0425 & N/A    &  N/A   &  N/A   & N/A  \\\hline
\multirow{2}{*}{2} & MSE (Sim-5-sec)  & N/A    & N/A    & N/A    & 0.2456 & 0.1162 & N/A \\
                   & MSE (Real-2-sec) & N/A    & N/A    & N/A    & 1.0729 & 0.1683 & N/A \\\hline
\multirow{2}{*}{3} & MSE (Sim-5-sec)  & 0.5229 & 0.2237 & N/A    & 0.2442 & 0.1259 & N/A  \\
                   & MSE (Real-2-sec) & 1.0657 & 0.2340 & N/A    & 1.0682 & 0.1468 & N/A  \\\hline
\multirow{2}{*}{4} & MSE (Sim-5-sec)  & 0.0941 & 0.9204 & 0.2800 & N/A    & N/A    &  N/A   \\
                   & MSE (Real-2-sec) & 0.1020 & 0.8102 & 0.3734 & N/A    & N/A    &  N/A  \\\hline
\multirow{2}{*}{5} & MSE (Sim-5-sec)  & N/A    & N/A    & N/A    & 0.2834 & 0.7982 & 0.5460  \\
                   & MSE (Real-2-sec) & N/A    & N/A    & N/A    & 1.0699 & 0.6459 & 0.5795  \\\hline
\multirow{2}{*}{6} & MSE (Sim-5-sec)  & 0.6321 & 0.9139 & 0.5245 & 0.3677 & 0.8661 & 0.5331 \\
                   & MSE (Real-2-sec) & 1.0846 & 0.7509 & 0.5353 & 1.0868 & 0.6039 & 0.5984  \\\hline
\end{tabular}
}
\caption{
Summary of Model Performance for each error state for scenarios 1-6 of the sim-5-sec and the real-2-sec datasets. The network is able to learn many of the errors for both simulated and real data ($MSE < 1$).
}
\label{sim_real_results}
\end{table}
%End- table 

%\begin{figure}[htb]
%    \centering
%    \includegraphics[width = 5 cm]{Figures/data4Error.png}
%    \includegraphics[width = 5 cm]{Figures/data1Error.png}
%    \caption{Distribution of Error States before (blue line) and after (histogram) adjusting to the neural network's error estimation for the real-2-sec data set. (Top) N Pos, E Pos and D Pos are the errors present in  dataset \# 4. (Bottom) N Pos and E Pos are the only errors present in  dataset \# 1. In both cases, the error distributions are more concentrated around zero, indicating that the network has improved our ability to estimate initial navigation errors. Furthermore, no outliers have been introduced by the network.}
%    \label{fig:Data4Errors}
%\end{figure}

\begin{figure}
    \centering
    \includegraphics[width=0.28\textwidth]{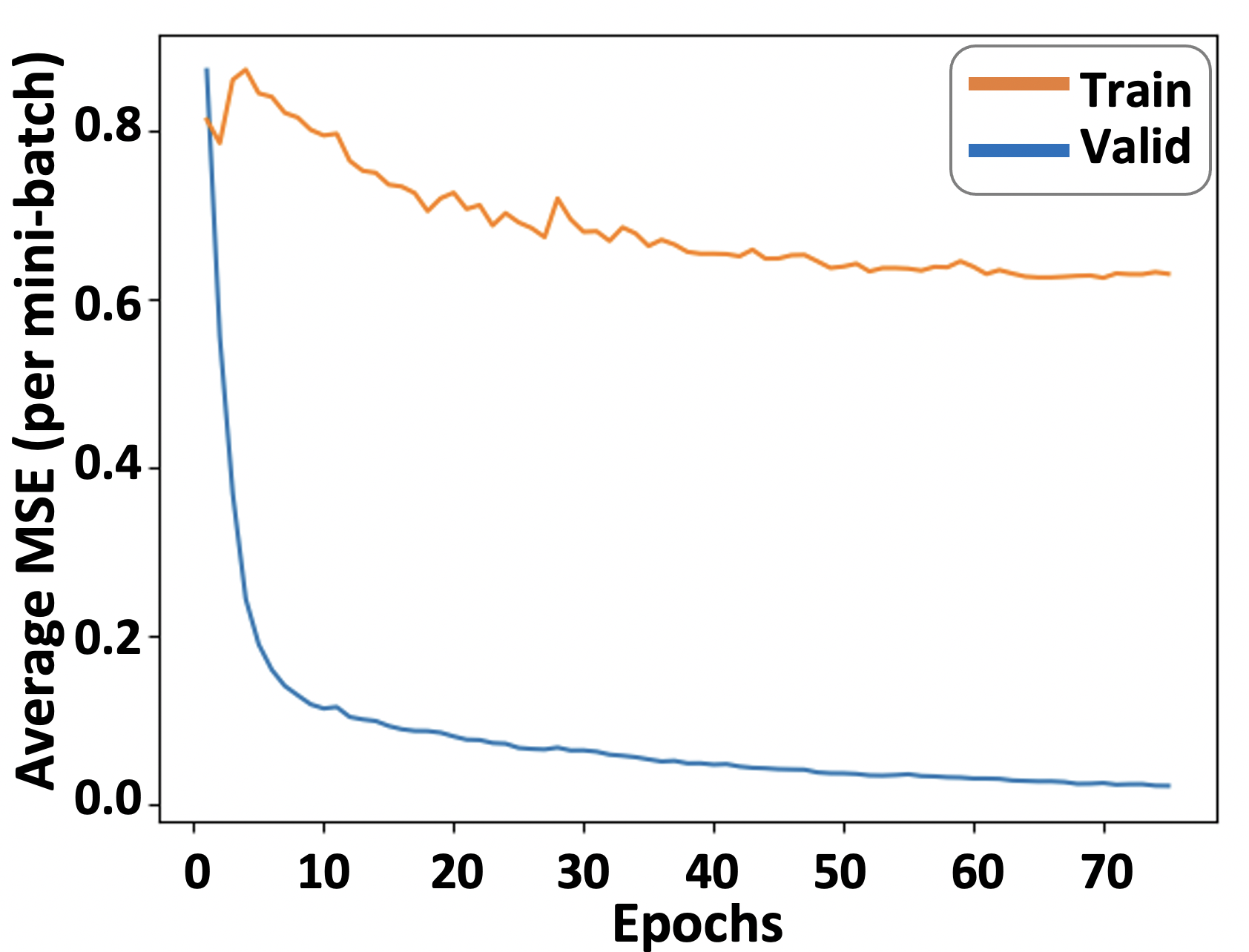}
    \caption{Training and validation MSE as a function of training epoch for scenario 2 for the real-2-sec dataset. There is a large gap between the training and validation error, suggesting the network may be overfitting. Compare with \autoref{fig:Accuracy_training_Data1} where little overfitting is occurring in scenario 1.}
    \label{fig:Accuracy_training_Data2}
\end{figure}

For some of these errors, the network may also be overfitting. \autoref{fig:Accuracy_training_Data2} shows the training and validation MSE as a function of training epoch for scenario 2. Here, we see that the average training error is very low, indicating that the network is able to accurately predict both errors on the training set. However, the validation MSE is considerably higher, suggesting that overfitting is occurring. Increasing the amount of training data, especially adding more targets, will likely reduce the tendency to overfit.
 
%Adding the accuracy plot Dataset 2

As observed in scenario 2 of the real-2-sec dataset, the network was not able to characterize blur distortions for the 2-second synthetic aperture despite the lack of error ambiguity.  \autoref{real_data_10sec_results} summarizes the network's performance in this scenario for a 2-second aperture, as compared to a 10-second aperture. In the latter case, the MSE for the North velocity error decreased by more than 25\%, suggesting that the neural network successfully characterized the effect of blurs for the larger apertures. Notably, the neural network also improves its estimation of the East velocity error by a factor of 2. These results identify a trend of improved performance on real datasets with increasing aperture length.

\begin{table}[!ht]
\centering
\scalebox{0.8}{
\begin{tabular}{|c|l|c|}\hline
                    Dataset, Scenario 2 &  AT Vel &  CT Vel \\\hline
           MSE (Real-2-sec)  &    1.0729    &    0.1683    \\\hline
          MSE (Real-10-sec)  &    0.7895    &    0.0812    \\\hline
MSE (Transfer Learning - 2 sec)  &    1.0429    &    0.0849    \\\hline
\end{tabular}
}
\caption{
Summary of Model Performance for the error states present in scenario 2. This table shows: 1) the results of applying the proposed model on the real-2-sec and real-10-sec data; 2) the results of pretraining on simulated data and training on real-2-sec data. Going from 2 seconds to 10 seconds improves the performance. Improvements from transfer learning are negligible.
}
\label{real_data_10sec_results}
\end{table}

We also applied transfer learning, where we trained the network on the sim-5-sec dataset first, then trained on the real-2-sec dataset. \autoref{real_data_10sec_results} shows the performance for scenario 2. We see that transfer learning resulted in negligible improvement in both the AT and CT velocity MSE. Similar results were obtained in  other scenarios.

\section{Conclusion}

We used a convolutional neural network to estimate position and velocity errors at the beginning of a SAR data collection period, by comparison of a distorted SAR image to an \textit{a priori} SAR reference image.  Performance was assessed on both simulated and real SAR data. In general, the network performs well in the absence of ambiguous error sources, reducing the MSE of the active navigation errors.  As the number of error sources increases, the performance degrades due to the inability to separate errors which result in the same image distortion.  In the case of the real data, however, the network successfully separated CT shifts caused by CT and D position errors.  This exciting result suggest that the network is able to identify subtleties in the image distortions that are not readily apparent to human eyes.  Another key finding in the testing of the network on real data, is sensitivity of estimation performance on the length of the synthetic aperture.  As the length increases, the network successfully characterizes blurring in real images, and appropriately attributes it to the corresponding error source.

Future work includes further investigation of the effect of the aperture length on the network's ability to learn. Furthermore, the sensitivity of distortions to the vehicle/target geometry may be exploited to reduce ambiguities and improve performance. Future approaches may, therefore, augment the training data with information about the viewing geometry, including the estimated vehicle position at the beginning/end of the synthetic aperture, the location of known targets, or the time history of barometric altimeter measurements.  Finally, including more training data in future iterations will likely help mitigate overfitting.

\bibliographystyle{IEEEbib}
\bibliography{references}

\end{document}